\pdfoutput=1

\documentclass[11pt]{article}

\usepackage[]{acl}

\usepackage{times}
\usepackage{latexsym}
\usepackage{amsfonts}

\usepackage[T1]{fontenc}

\usepackage[utf8]{inputenc}

\usepackage{microtype}

\usepackage{inconsolata}

\usepackage{subcaption}
\usepackage{booktabs}
\usepackage{graphicx}
\usepackage{enumitem}
\usepackage{amsmath} 
\usepackage{listings}
\usepackage{adjustbox}

\usepackage{colortbl}

\usepackage{bm}

\usepackage{multirow}
\usepackage{xcolor}
\usepackage{hyperref}
\usepackage{xspace}

\addtocontents{toc}{\protect\setcounter{tocdepth}{-1}}

\usepackage{pifont}   
\usepackage{booktabs}

\usepackage{amsfonts}  
\usepackage{array}   

\usepackage{graphicx}
\usepackage{amsmath}
\usepackage{amssymb}
\usepackage{booktabs}
\usepackage{amsfonts}
\usepackage{algorithmic}
\usepackage{multicol,multirow}
\usepackage{makecell}

\definecolor{darkgreen}{RGB}{50,100,0}
\definecolor{darkred}{RGB}{200, 0, 0}
\definecolor{lightred}{RGB}{250, 200, 200}
\definecolor{lightblue}{RGB}{200, 200, 250}

\newcommand{\cmark}{\textcolor{darkgreen}{\ding{51}}} %
\newcommand{\xmark}{\textcolor{darkred}{\ding{55}}} %

\definecolor{pink}{rgb}{0.858, 0.188, 0.478}  
\hypersetup{colorlinks=true, urlcolor=pink}

\newcommand{\model}[1]{\textsc{LLM-Neo}} %

\lstdefinestyle{python}{
    language=Python,
    basicstyle=\fontsize{7}{9.5}\ttfamily,
    keywordstyle=\color{blue},
    commentstyle=\color{gray},
    stringstyle=\color{black},
    showstringspaces=false,
    breaklines=true,
    breakindent=0pt,
    breakatwhitespace=false,
    escapeinside={(*@}{@*)}
}

\definecolor{gg}{HTML}{e2f0cb}

\newcommand*{\affmark}[1][*]{\textsuperscript{#1}}
\usepackage{fdsymbol}

\title{\textsc{Llm-Neo}: Parameter Efficient Knowledge Distillation \\ for Large Language Models}

\author{
Runming Yang\affmark[$\clubsuit$]\thanks{Equal contributions. Work was done when Runming was interning at Tencent.} \
Taiqiang Wu\affmark[$\diamondsuit$]\footnotemark[1] \ 
Jiahao Wang\affmark[$\diamondsuit$] \
Pengfei Hu\affmark[$\spadesuit$] \\ \textbf{Yik-Chung Wu}\affmark[$\diamondsuit$] \ \textbf{Ngai Wong}\affmark[$\diamondsuit$] \ 
\textbf{Yujiu Yang}\affmark[$\clubsuit$]
\\
\affmark[$\clubsuit$]Tsinghua University \
\affmark[$\diamondsuit$]The University of Hong Kong \ 
\affmark[$\spadesuit$]Tencent
\\
{\tt yrm22@mails.tsinghua.edu.cn} \ {\tt yang.yujiu@sz.tsinghua.edu.cn}
}

\begin{document}
\maketitle
\begin{abstract}
Knowledge distillation (KD) has been a predominant method for compressing Large Language Models (LLMs).
In this paper, we first revisit KD and Low-Rank Adaption (LoRA) and demonstrate that they follow the same paradigm.
Inspired by this observation, we propose a parameter-efficient knowledge distillation method, \textsc{Llm-Neo}, which integrates LoRA into KD to improve the efficiency of knowledge transfer.
After that, we summarize some valuable guidelines for the hyperparameters in \textsc{Llm-Neo}.
Experimental results on compressing Llama 2 and Llama 3.2 show that \textsc{Llm-Neo} outperforms various baselines.
Further analysis demonstrates the robustness of the proposed \textsc{Llm-Neo} on variants of LoRA.
The code and trained models are available at \href{https://github.com/yang3121099/LLM-Neo}{Github}.
\end{abstract}

\section{Introduction}

Knowledge distillation (KD)~\cite{hinton2015distilling} is a predominant method to compress large language models (LLMs) \cite{gpt}.
The key insight is to train a compact student model by mimicking the behaviors of the teacher model.
One mainstreaming way is to align the logits ~\cite{wu2024rethinking}, and thus transfer the knowledge from the teacher model to the student model.

Parameter-Efficient Fine-Tuning~(PEFT)~\cite{lester2021power,he2021towards} is another commonly used technique for LLM efficiency~\cite{han2024parameter}. 
Among various PEFT methods, the Low-Rank Adaption~(LoRA)~\cite{hu2021lora} has gained increasing popularity since it does not introduce any additional parameters for inference.
During training, LoRA updates a mergeable low-rank branch instead of updating the original full parameters. 
Therefore, LoRA can efficiently transfer the knowledge contained in the training examples to the trained models.

%%%%%%%%%
\begin{figure}[!t]
\centering
\includegraphics[width=0.42\textwidth]{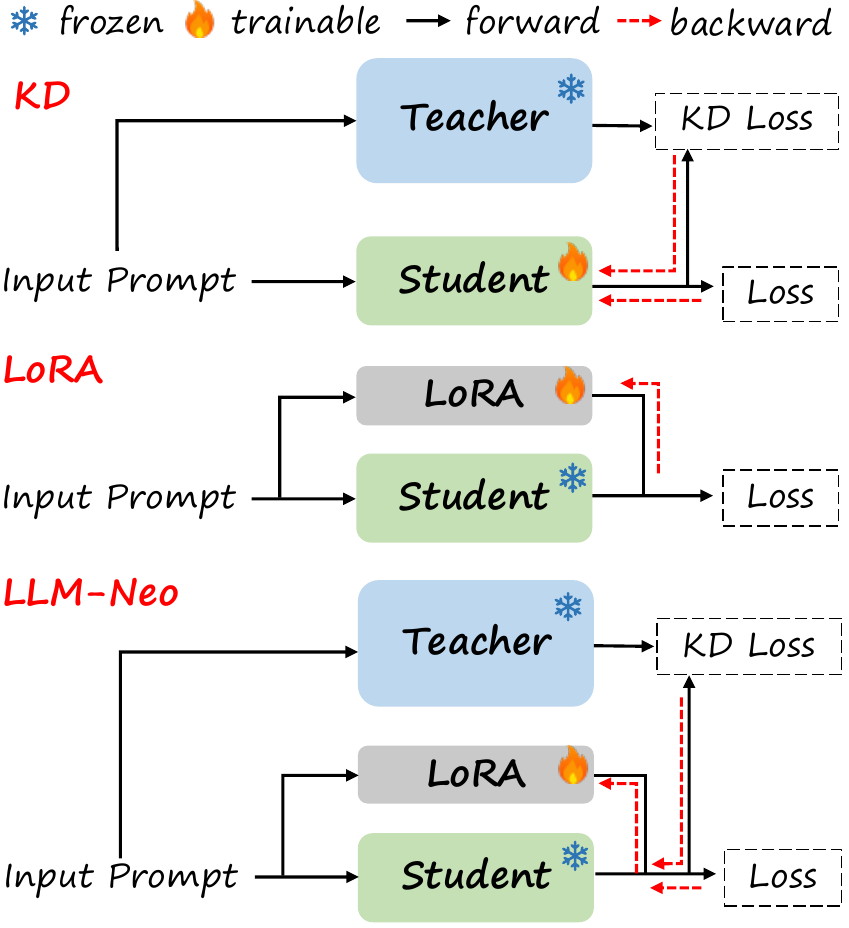} 
\caption{
    Illustration of different knowledge transfer pipelines (KD, LoRA, and \textsc{Llm-Neo}).
    The proposed \textsc{Llm-Neo} pipeline combines the benefits of both the KD and LoRA approaches, i.e., distilling knowledge from the teacher and low-rank branch efficiency.
}
\label{fig:intro}
\end{figure}
%%%%%%%%%

In this paper, we first show that KD and LoRA follow the \textbf{same paradigm}, i.e., aiming at transferring knowledge while the sources differ~\cite{wan2023efficient}.
Moreover, LoRA transfers the knowledge efficiently via the low-rank branch, while KD methods update the full parameters and typically cost much more resources.
We thus ask: \textit{can we combine KD and LoRA to improve the efficiency of knowledge transfer from the teacher model}?

To this end, we propose a novel \textsc{Llm-Neo} framework that integrates LoRA into KD to achieve parameter-efficient knowledge distillation. 
Specifically, as shown in Figure~\ref{fig:intro}, we follow the idea of LoRA to introduce a low-rank branch in the student model, aiming to inherit the knowledge from the teacher model. 
We first perform comprehensive analysis and derive valuable \textit{guidelines} for the design of the proposed \textsc{Llm-Neo}.
Experimental results on compressing Llama 2~\cite{touvron2023llama2} and Llama 3.1~\cite{llama3modelcard} demonstrate the effectiveness and efficiency of \textsc{Llm-Neo}. 
Moreover, further analysis shows the robustness of \textsc{Llm-Neo} towards LoRA variants~\cite{wu2024mixture}.
Our contributions can be concluded as follows: 
% \todo{contributes,done}

\begin{itemize}
    \item We show that KD and LoRA share a unified paradigm, and thus propose a novel method, \textsc{Llm-Neo}, to improve the efficiency of knowledge transfer from the teacher model.
    \item We summarize three key guidelines for \textsc{Llm-Neo}, namely: i) typically a higher rank makes better accuracy while harder to converge;
    ii) a learning rate about 2e-4 works well;
    iii) a larger rank requires a lower learning rate. 
    \item We perform extensive experiments on the Llama 2 and Llama 3 models to demonstrate the effectiveness of \textsc{Llm-Neo}. 
    Further analyses highlight the effectiveness of improving model performance and scalability. 
\end{itemize}

\section{Methodology}
\subsection{Unified Paradigm for KD and LoRA}

We can decompose the model parameters $W_t$ at step $t$ as: $W_t = W_0 + \Delta W_t$, where $W_0$ denotes the vanilla pretrained weights and $\Delta W_t$ for the weight difference.
Thus, the unified paradigm for updating parameters ($W_t\rightarrow W_{t+1}$) is:
\begin{equation}
\underbrace{W_{t+1}}_{\text{Updated}} \leftarrow { W_t} - \eta \cdot \frac{\partial {\mathcal{L}}_{\textcolor{red}{\mathcal{D}}}}{\partial (W_0 + \textcolor{red}{f}(\Delta W_t))}.
\label{eq: unified_format}
\end{equation}
In this equation, two key components emerge: 1) loss function on the \textit{information source} \textcolor{red}{$\mathcal{D}$} and 2) \textit{mapping function} \textcolor{red}{$f(\cdot)$}. 

\paragraph{KD.} KD aims to transfer the knowledge from teacher to student, which can be formulated as:
\begin{equation}
W_{t+1} \leftarrow W_t - \eta \cdot \frac{\partial \mathcal{L}_{KD}}{\partial (W_0 + \Delta W_t)},
\end{equation}
\begin{equation}\label{eq:kd-loss}
\mathcal{L}_{KD} = \alpha \cdot \mathcal{L}_{CE}(y, z_s) + (1 - \alpha) \cdot \mathcal{L}_{KL}(z_t, z_s),
\end{equation}
where $\alpha$ is a hyperparameter to combine guidance information from ground truth and teacher, and $z_t$ is the output logits of the teacher.
In the unified view, we have $\textcolor{red}{f}(\Delta W_t) = \Delta W_t$ and $\mathcal{\textcolor{red}{D}} = \{\text{Ground Truth}, \text{Teacher}\}$.

\begin{table}[t]
\centering
\small

\resizebox{0.5\textwidth}{!}{
\begin{tabular}{>{\raggedright}p{15mm}c >{\centering}m{15mm} c c}
\toprule
\multirow{2}{*}{\textbf{Method}} & 
\multirow{2}{*}{$\bm{\mathcal{D}^\dagger}$} & 
\multirow{2}{*}{$\bm{f(x)}$} & 
\textbf{Teacher's} & \textbf{Parameter} \\
&&&\textbf{Guidance} & \textbf{Efficiency} \\
\midrule
LoRA 
& GT 
& $x - W_0$ 
& \xmark 
& \cmark \\ [0.25em]

KD 
& GT + $\mathcal{T}$
& $x$ 
& \cmark 
& \xmark \\ [0.25em]

\rowcolor{gray!10}
\textsc{Llm-Neo} 
& GT + $\mathcal{T}$
& $x - W_0$ 
& \bm{\cmark} 
& \bm{\cmark} \\ [0.25em]
\bottomrule
\end{tabular}
}
\caption{Comparison of LoRA, KD, and \textsc{Llm-Neo}. 
GT denotes the ground truth and $\mathcal{T}$ for the teacher.
We highlight the unified advantages of \textsc{Llm-Neo} in combining teacher guidance with parameter-efficient low-rank adaptation.}
\label{tab:method_compare}
\vspace*{0.35em}
\vspace{-2ex}
\end{table}

\paragraph{LoRA.} LoRA constrains updates to low-rank matrices.
The optimization process is defined as:
\begin{equation}
 W_{t+1} \leftarrow  W_t - \eta \cdot \frac{\partial \mathcal{L}_{CE}(z_t, y)}{\partial (W_0+\Delta W_t-W_0)},
\end{equation}
where $y$ is the ground truth label, $z_s$ is the logits of the student, and $\Delta W_t := A_tB_t^\top$.
We optimize $A_t$ and $B_t$ while keeping $W_0$ frozen.
In the unified view, we have $\textcolor{red}{f}(\Delta W_t) = \Delta{W}_t-W_0$ and $\mathcal{\textcolor{red}{D}} = \{\text{Ground Truth}\}$.
\paragraph{Summary.} Both LoRA and KD follow the paradigm defined in Equation \ref{eq: unified_format}, while the data source $\mathcal{D}$ and mapping function $f(\cdot)$ differ.

\subsection{\textsc{Llm-Neo} Method}
Motivated by the observation, we propose the \textsc{Llm-Neo} framework synthesizes LoRA's structural constraints with KD's multi-source guidance. 
Specifically, we have $\textcolor{red}{f}(\Delta W_t) = \Delta{W}_t-W_0$ following LoRA and $\mathcal{\textcolor{red}{D}} = \{\text{Ground Truth}, \text{Teacher}\}$ following KD.
\textsc{Llm-Neo} updates the low-rank branch as follows:
\begin{equation}
W_{t+1} \leftarrow W_t - \eta \cdot \frac{\partial\mathcal{L}_{KD}}{\partial\Delta W_t}.
\end{equation}

Table \ref{tab:method_compare} indicates the comparison among vanilla KD, LoRA, and proposed \textsc{Llm-Neo}.
By this design, \textsc{Llm-Neo} can enjoy both teacher guidance and parameter efficiency.

\section{Experiments}

\begin{table*}[!t]
\centering

\resizebox{\textwidth}{!}{
    \begin{tabular}{l|cc|cccc|cc|cccc}
        \toprule
        \multirow{2}{*}{\textbf{Metric}} & \multicolumn{6}{c|}{Llama 3.1-8B $\longrightarrow$  Llama 3.2-1B} & \multicolumn{6}{c}{Llama 2-7B $\longrightarrow$ TinyLlama-1.1B} \\
        \cmidrule(lr){2-7} \cmidrule(lr){8-13}
        & \textbf{Teacher} & \textbf{Student} & \textbf{SFT} & \textbf{LoRA} & \textbf{KD} & \cellcolor{gg}{\textbf{\textsc{Llm-Neo}}} & \textbf{Teacher} & \textbf{Student} & \textbf{SFT} & \textbf{LoRA} & \textbf{KD} & \cellcolor{gg}{\textbf{\textsc{Llm-Neo}}} \\
        \midrule
        Mem   & - & - & 63G & 68G & 231G & 177G & - & - & 66G & 42G & 167G & 136G \\
        Time  & - & - & 10min & 7min & 25min & 20min & - & - & 13min & 12min & 26min & 25min \\
        \midrule
        ARC-e  & 81.90 & 68.52 & 67.72 & 68.39 & 69.15 & 69.11 & 76.73 & 60.27 & 60.61 & 60.35 & 61.49 & 61.24  \\
        ASDiv  &88.70 & 67.40  & 67.80 & 67.40 & 67.90 & 69.00 &61.70  & 18.10 &21.60 & 20.80& 21.70  &21.30 \\
        HellaS. & 59.10 & 45.07 & 45.20 & 45.30 & 45.29 & 45.38 & 56.47 & 44.99 & 46.82 & 46.86 & 46.73 & 46.72  \\
        PIQA   & 80.09 & 73.88 & 74.70 & 74.43 & 74.97 & 74.65 & 78.35 & 74.34 & 72.69 & 72.52 & 73.34 & 73.23  \\
        WinoG. & 73.72 & 59.27 & 61.96 & 60.69 & 60.54 & 61.25 & 71.03 & 58.72 & 60.22 & 59.83 & 59.91 & 60.54  \\
        \textbf{Avg.}   &76.70  & 62.83 & 63.48 & 63.24 & 63.57 & \cellcolor{gg}{\textbf{63.88}} &  68.86 &51.28 &52.39 & 52.07& 52.63& \cellcolor{gg}{\textbf{52.60}} \\
        \bottomrule
    \end{tabular}
}
\caption{Comparison of SFT, LoRA, KD, and \textsc{Llm-Neo} on 5 benchmarks. 
We perform KD from Llama 3.1-8B to Llama 3.2-1B, and from Llama 2-7B to TinyLlama-1.1B. 
\textsc{Llm-Neo} achieves the best average performance, with superior memory and time efficiency compared to KD.}
\label{tab:combined_comparison}
\end{table*}

\subsection{Experimental Implement}

For the training data, we employ the BAAI Infinity-Instruct dataset
\citep{InfinityInstruct2024} and randomly sample 10,000 samples as fine-tuning data (around 5M tokens).

For evaluation, we employ several popular reasoning benchmarks, including MMLU \cite{DBLP:journals/corr/HuangBZ23}, CMMLU~\cite{li2023cmmlu}, C-Eval~\cite{DBLP:conf/iclr/HendrycksBBZMSS21}, PIQA~\cite{Bisk2020piqa}, HellaSwag~\cite{zellers2019hellaswag}, WinoGrande~\cite{sakaguchi2019winogrande}, ARC-easy~\cite{clark2018think}, ARC-challenge~\cite{clark2018think}, and OpenbookQA~\cite{OpenBookQA2018}) using the lm-evaluation-harness package~\cite{eval-harness}.
Besides, we employ several benchmarks for the math ability, including ASDiv~\cite{miao2021diverse}, GSM8K~\cite{cobbe2021gsm8k}, SVAMP~\cite{patel2021nlp}, MAWPS~\cite{koncel2016mawps}, and MathQA~\cite{amini2019mathqa}.

We set the maximum input length to 512, the batch size to 4, and the gradient accumulation steps to 16 for all experiments.
It takes about an hour to train on 6 Nvidia A100 40G GPUs for 2 epochs.

%%%%%%%%%%%%%%%%

\begin{figure}[!t]
\centering
\includegraphics[width=0.48\textwidth]{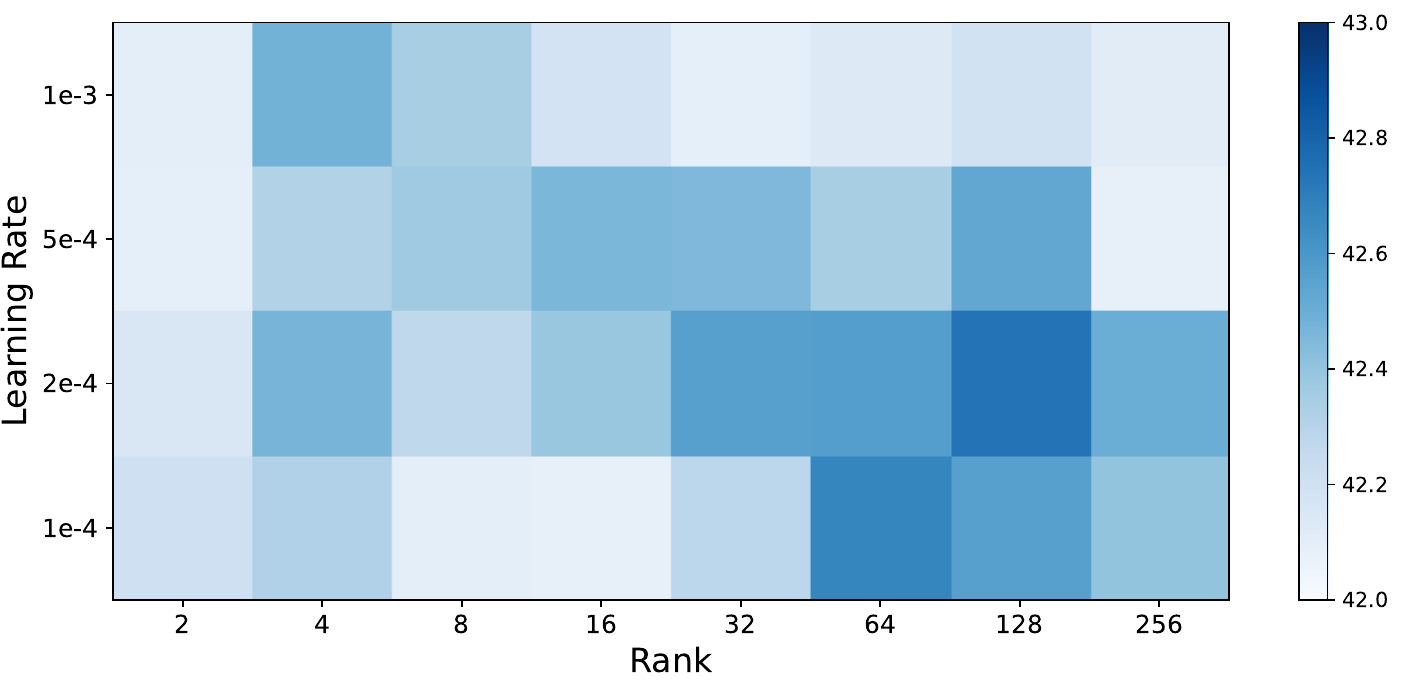} 
\caption{
Grid search results from Llama 2 to TinyLlama for rank (2 to 256).
The score matrix shows the average of 10 reasoning metrics, with darker colors indicating better performance. 
}
\vspace{-14pt}

\label{fig:Preliminary}
\end{figure}
%%%%%%%%%
%%%%%%%%%%%%%%%%

\subsection{Guidelines}
\label{sec: setting for rank and lr}

We first distill the Llama 2-7B \cite{touvron2023llama2} to the TinyLlama-1.1B \cite{zhang2024tinyllama} on the 100,000 training samples.
We conduct a grid search for the learning rates in \{1e-4, 2e-4, 5e-4, and 1e-3\} and rank $r$ in \{2, 4, 6, 8, 16, 32, 48, 128, and 256\}.
Figure \ref{fig:Preliminary} shows the average evaluation results on the reasoning benchmarks.
We can conclude the following \textit{guidelines}:
\begin{itemize}
    \item Typically, the larger the rank, the better.
    Different with fine-tuning, KD shows a significant trends that larger rank would bring better results.
    \item A learning rate close to 2e-4 works well for LoRA, and the larger one (such as 1e-3) would lead to performance decrease.
    \item Considering the relationship between rank and learning rate, 
    a larger rank requires a lower learning rate.
\end{itemize}

\subsection{Main Results}

Based on the guidelines, we set the rank as 128 and learning rate as 2e-4.
For Llama 2 series experiments, we employ Llama 2-7B-chat as the teacher model and TinyLlama-1.1B as the student model. 
For Llama 3 series experiments, we employ Llama 3.1-8B-Instruct as the teacher and Llama 3.2-1B-Instruct as the student model. 
% \todo{not pruned}
Moreover, we also conduct knowledge distillation on Llama 3-pruned-1B, which is pruned from the Llama 3.1 8B model \citep{llmpruner,kim2024shortened}, and on Minitron-4B \cite{turuvekere2024llm,bansal2024smaller}.
The minimum learning rate is set to 1e-5 for LoRA-base experiments and 1e-6 for SFT and KD experiments.
We employ the 50M tokens dataset to perform the distillation.

Table \ref{tab:combined_comparison} shows the results on Llama 3.2 1B and TinyLlama 1B.
Please refer to Appendix \ref{appendix: Llama-3-pruned-1B} and \ref{appendix: minitron} for the results on Llama 3-pruned-1B and Minitron-4B.
Overall, the results demonstrate that our \textsc{Llm-Neo} approach outperforms KD in memory efficiency and training time, while also surpassing SFT and standard LoRA methods. 
Specifically, \textsc{Llm-Neo} gets an average score of 63.88 regarding the Llama 3.2 1B, which is 0.64 higher than the LoRA and 0.31 higher than the KD.
Considering the Llama 2, \textsc{Llm-Neo} also outperforms the baselines.
Meanwhile, \textsc{Llm-Neo} can save up to 25\% GPU memory and training time, demonstrating the efficiency of knowledge transferring.
Experiments on the Llama 2 and Llama 3 series further prove the robustness of our proposed \textsc{Llm-Neo}.

% \vspace{-10em}
%%%%%%%%%
\begin{figure}[]
\centering
\includegraphics[width=0.45\textwidth]{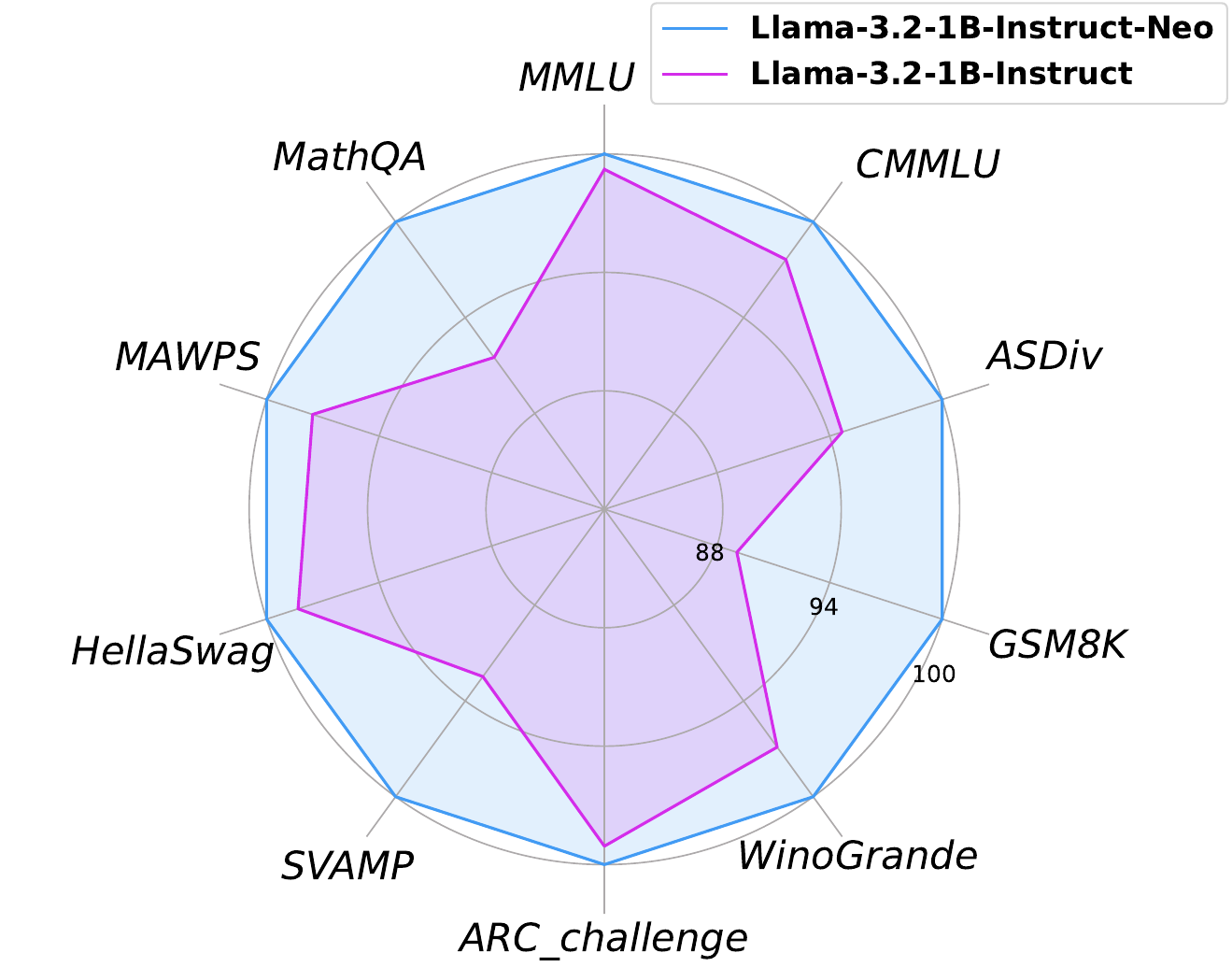} 
\caption{
% \todo{caption}.
Normalized performance on 10 benchmarks for Llama 3.2-1B Instruct model before and after distillation via \textsc{Llm-Neo}.
}
\vspace{-1em}
\label{fig:rader}
\end{figure}
%%%%%%%%%

\paragraph{More benchmarks.} Besides the 5 benchmarks shown in Table \ref{tab:combined_comparison}, we also visualize results on more benchmarks.
As shown in Figure \ref{fig:rader}, we can conclude that \textsc{Llm-Neo} shows effectiveness and robustness on all 10 benchmarks.
particularly, \textsc{Llm-Neo} shows strong ability on the math tasks, such as SVAMP and GSM8K datasets.

\section{Extensive Analysis}

We further perform extensive analysis on the setting from Llama 3.1 8B to Llama 3-pruned-1B.

\subsection{Strengthen with LoRA variants}
\begin{figure}[!t]
\centering
\includegraphics[width=0.45\textwidth]{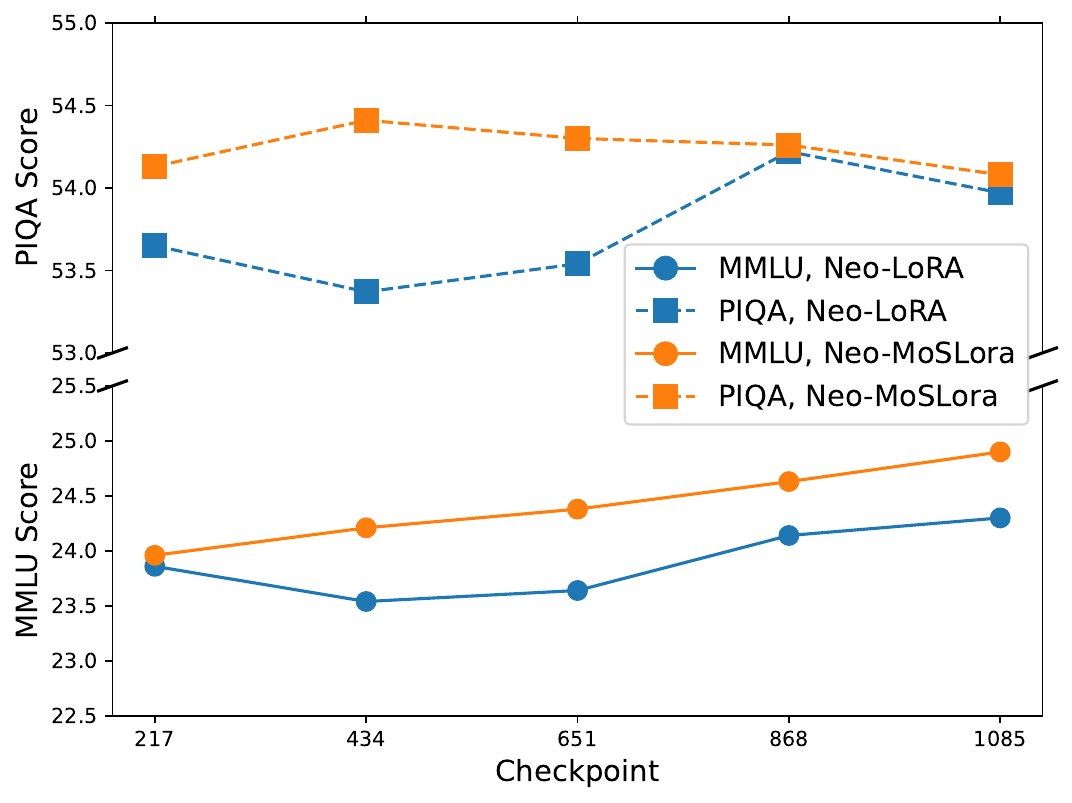} 
\caption{
Comparison of vanilla \textsc{Llm-Neo} and \textsc{Llm-Neo}-MoSLoRA on MMLU and PIQA.
}
\label{fig:moslora}
\end{figure}

To evaluate the robustness of \textsc{Llm-Neo} towards LoRA variants, we try the latest variant of MoSLoRA \cite{wu2024mixture}, which improves LoRA via mixing the subspaces. 
As shown in Figure \ref{fig:moslora}, \textsc{Llm-Neo}-MoSLoRA gets better performance than vanilla LoRA consistently,  which demonstrates the robustness of \textsc{Llm-Neo}.

\subsection{Data scaling law in \textsc{Llm-Neo}}
\begin{figure}[t]
\centering
\includegraphics[width=0.45\textwidth]{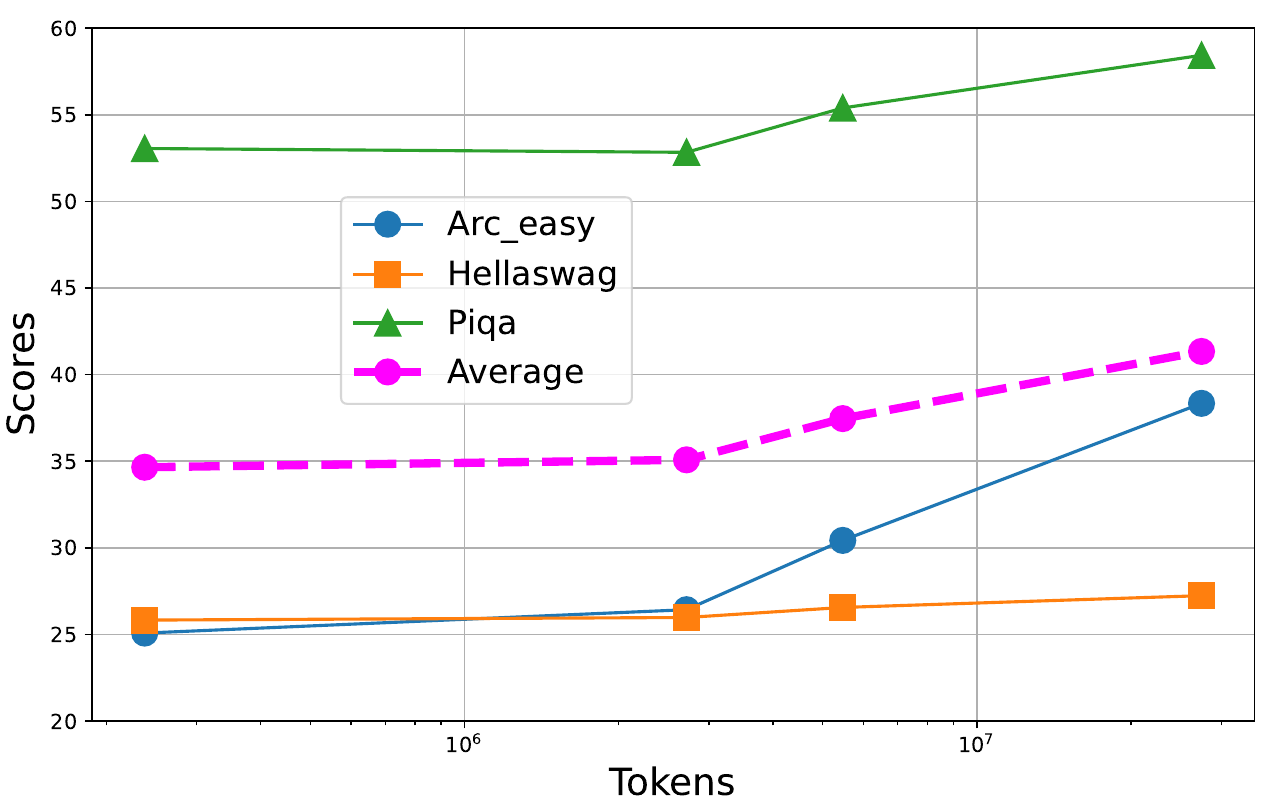} 
\caption{
Performance with more tokens range increased from $10^6$ to $10^8$. 
% The performance would be better with more training tokens.
}
\label{fig:scaling law}
\end{figure}

Following the Llama3 report, we also scale the dataset and apply \textsc{Llm-Neo} to larger datasets progressively, including 100K, 200K, 500K, and 1M training samples. 
As shown in Figure \ref{fig:scaling law}, the results indicate a consistent improvement in performance as the data size increased, highlighting its scalability and robustness to larger datasets.

\subsection{Compatibility with More Memory Optimizations}
For more memory optimization strategies, we also conduct \textsc{Llm-Neo} on ZeRO-1 and ZeRO-2.
Experimental results demonstrate its robustness towards various optimization strategies.
More details can be found at Appendix \ref{deepspeed_apex}.

%%%%%%%%%
%%%%%%%%%

\section{Conclusion}
In this work, we propose a novel \textsc{Llm-Neo} framework, aiming to efficiently transfer knowledge from a large language model (LLM) teacher to a compact student.
We first revisit the KD and LoRA, and provide a unified paradigm.
Therefore, we explore the strategy combining LoRA and KD to enhance the efficiency of knowledge transfer.
We first summarize some guidelines, i.e., the larger rank around 128 is more suitable.
Experimental results on compressing Llama 2 and Llama 3.1 show that \textsc{Llm-Neo} outperforms various baselines.
For future work, we would like to explore the performance of \textsc{Llm-Neo} on more tasks and models with different structures.

\clearpage

\section{Limitations }
In this study, we provide a unified framework for knowledge acquisition through KD and LoRA, and thus propose \textsc{Llm-Neo} as a cross-strategy that combines the advantages of both methods, which improves efficiency while maintaining capability.
To evaluate \textsc{Llm-Neo}, we perform experiments on the Llama series and TinyLlama and report results on various benchmarks.
However, there are lots of LoRA variants and we only cover MoSLoRA.
Fortunately, it is easy to apply these LoRA variants on \textsc{Llm-Neo} framework.
We leave it for future work.
\section{Ethics Statement}

This work aims to improve the efficiency of knowledge distillation following the idea of LoRA and verified on various LLMs.
However, it also inherits the social risks of generative LLMs, such as gender and representation bias \citep{lucy-bamman-2021-gender}.
Fortunately, the proposed \textsc{Llm-Neo} can be applied to various LLMs and we encourage deploying the risk-free LLMs to reduce the potential ethical risks.

\bibliography{custom}

\clearpage

\appendix
\onecolumn

\begin{table*}[!h]
\centering
\resizebox{0.6\textwidth}{!}{
    \begin{tabular}{l|cc|cccc}
        \toprule
        \multirow{2}{*}{\textbf{Metric}} & \multicolumn{6}{c}{Llama 3.1-8B $\longrightarrow$  Llama 3-pruned-1B}  \\
        \cmidrule(lr){2-7} 
        & \textbf{Teacher} & \textbf{Student} & \textbf{SFT} & \textbf{LoRA} & \textbf{KD} & \cellcolor{gg}{\textbf{\textsc{Llm-Neo}}} \\
        \midrule
        Mem   & - & - & 63G & 68G & 231G & 177G \\
        Time  & - & - & 10min & 7min & 25min & 20min  \\
        \midrule
        ARC-e  & 81.90 & 28.07 & 30.39 & 32.95 & 34.85 & 34.89  \\
        CEVAL  & 53.94 & 25.33 & 25.63 & 24.15 & 23.63 & 24.00  \\
        HellaS. & 59.10 & 26.00 & 26.67 & 26.67 & 27.08 & 27.14 \\
        PIQA   & 80.09 & 53.92 & 54.41 & 56.09 & 57.45 & 56.58  \\
        WinoG. & 73.72 & 50.43 & 51.38 & 51.85 & 52.64 & 52.64  \\
        \textbf{Avg.}   & 69.35 & 36.35 & 37.58 & 38.34 & 39.13 & \cellcolor{gg}{\textbf{39.21}} \\
        \bottomrule
    \end{tabular}
}

\caption{Comparison of SFT, LoRA, KD, and \textsc{Llm-Neo} on 5 benchmarks. 
The results from Llama 3.1-8B to Llama 3-pruned-1B and \textsc{Llm-Neo} achieves best average performance, with superior memory and time efficiency compared to the KD method.}
\label{appendix:A}
\end{table*}

\section{Results on Llama 3-pruned-1B}
\label{appendix: Llama-3-pruned-1B}

To demonstrate the advantages of \textsc{Llm-Neo} in practical applications, we employ Llama3.1-8B-Instruct as the teacher and the Llama 3-pruned-1B as the student model. 
The minimum learning rate is set to 1e-5 for LoRA-base experiments and 1e-6 for SFT and KD experiments.
We employ the 50M tokens dataset to perform the distillation.

As shown in Table \ref{appendix:A}, the results of the comparative analysis of SFT, LoRA, KD, and \textsc{Llm-Neo} demonstrate that our \textsc{Llm-Neo} approach outperforms KD in memory efficiency and training time, which can save about 25\% GPU memory and training time, while also surpassing SFT and standard LoRA methods in overall performance. 
Experiments on the Llama 3-pruned-1B further prove the robustness and the efficiency of knowledge transferring of our proposed \textsc{Llm-Neo}, and it can work well on pruned models which means \textsc{Llm-Neo} has orthogonality with other LLM lightweight methods.

\section{Results on Minitron-4B}
\label{appendix: minitron}
\begin{figure}[!h]
\centering
\includegraphics[width=0.45\textwidth]{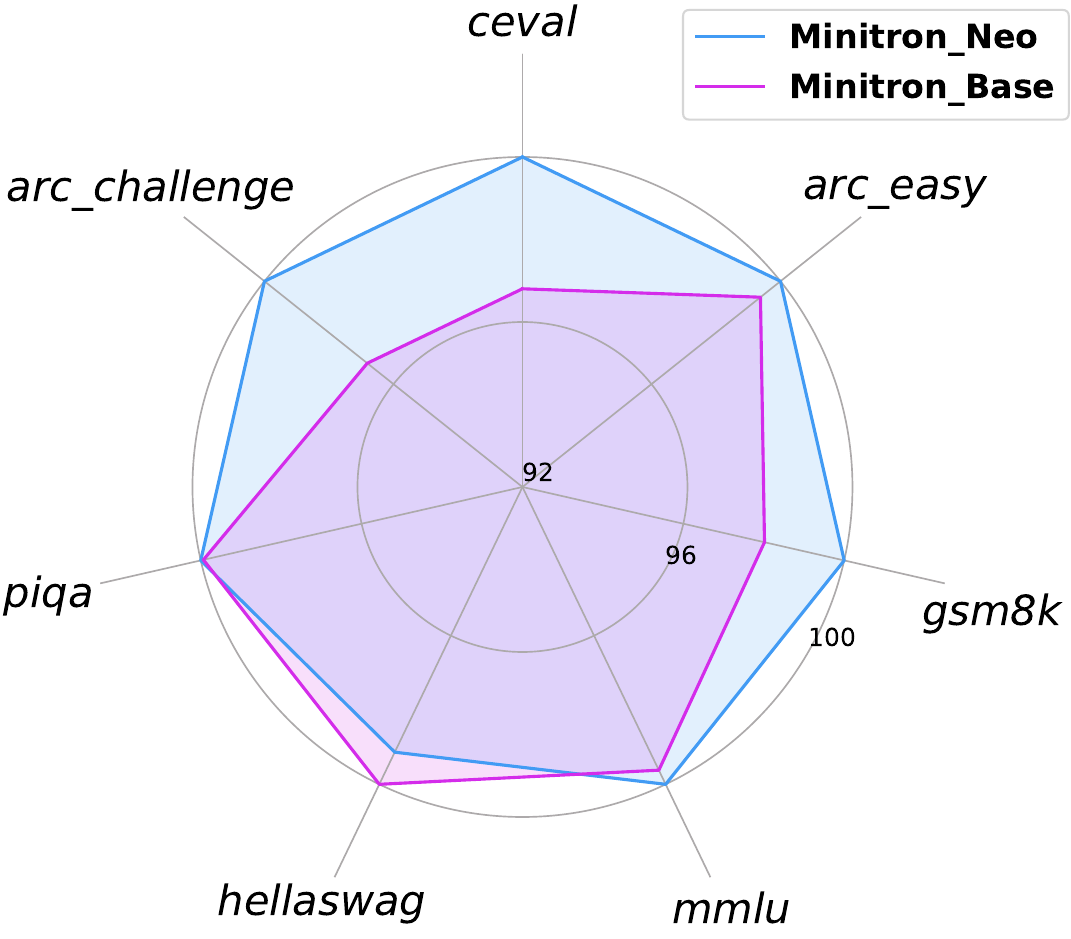} 
\caption{
Normalized performance on 10 ability dimensions for Minitron-4B-Depth-Base model finetuned on 50M tokens using the LLM-Neo method. The chart illustrates enhancements across multiple evaluation metrics, with the expanded area indicating overall performance gains.
}
\label{fig:minitron}
\end{figure}

We further perform distillation from Llama 3.1 to the Nvidia Minitron-4B \cite{turuvekere2024llm,bansal2024smaller} using the 50M tokens.
All the experiments are conducted on 6 A100 40G GPUs. 
Figure \ref{fig:minitron} shows the results on several benchmarks.
We can find that our proposed \textsc{Llm-Neo} performs well on the pruned Minitron 4B, demonstrating its robustness.
Specifically, \textsc{Llm-Neo} gets an average score of 54.37, which is 0.52 higher than the base model.
The trained weight has been available at HuggingFace.

% \vspace{-10em}
%%%%%%%%%
%%%%%%%%%

\section{Compatibility with More Memory Optimizations}
\label{deepspeed_apex}
To test the robustness of \textsc{Llm-Neo} combined with other memory optimizations, we explore the compatibility with existing LLM optimization techniques when distilling the Minitron 4B.
Specifically, we further conduct \textsc{Llm-Neo} with ZeRO-1 methods.
As shown in Table \ref{tab:deepspeed}, we can find that \textsc{Llm-Neo} works well with ZeRO1 and ZeRO2, highlighting the relationship between different ZeRO levels and their impact on performance metrics such as memory consumption and time efficiency.
Specifically, KD would be out-of-memory though we decrease the batch size to 1.

\begin{table}[htbp]
\centering

\resizebox{0.5\columnwidth}{!}{
    \begin{tabular}{l|c|c|c|cc}
        \toprule
        \multirow{2}{*}{\textbf{Metric}} & \multicolumn{3}{c|}{\textbf{Baseline Methods}} & \multicolumn{2}{c}{\textbf{\textsc{LLM-Neo}}} \\
        \cmidrule(lr){2-6}
        & \textbf{SFT} & \textbf{LoRA} & \textbf{KD} & \textbf{ZeRO1} & \textbf{ZeRO2} \\
        \midrule
        Mem   & 219G & 103G & OOM & 224G  & 212G  \\
        Time  & 35min & 22min & - & 40min & 37min  \\
        \bottomrule
    \end{tabular}
}
\caption{Memory usage and training time comparison across methods for the Minitron-4B-Width-Base model. Results show \textsc{LLM-Neo} compatibility with existing efficient methods.}
\label{tab:deepspeed}
\end{table}

\end{document}